\documentclass{article}

\usepackage{microtype}
\usepackage{booktabs} 
\usepackage{times}
\usepackage{epsfig}
\usepackage{graphicx}
\usepackage{subfigure}
\usepackage{amsmath}
\usepackage{amssymb}
\usepackage{xspace}
\usepackage{ mathrsfs }
\usepackage[utf8]{inputenc}
\usepackage{mathtools}
\usepackage{algorithm}
\usepackage{algorithmic}

\usepackage{balance}

\newcommand{\sP}{\mathsf{P}}

\usepackage{hyperref}
\usepackage{xcolor}

\usepackage[accepted]{icml2020}

\newcommand{\beginsupplement}{%
        \setcounter{table}{0}
        \renewcommand{\thetable}{A\arabic{table}}%
        \setcounter{figure}{0}
        \renewcommand{\thefigure}{A\arabic{figure}}%
        \setcounter{section}{0}
        \renewcommand{\thesubsection}{\Alph{subsection}}
        \renewcommand{\figurename}{Appendix, Figure}
     }

\icmltitlerunning{Meta-Meta Classification for One-Shot Learning }

\begin{document}

\twocolumn[
\icmltitle{Meta-Meta Classification for One-Shot Learning}




\begin{icmlauthorlist}
\icmlauthor{Arkabandhu Chowdhury}{rice}
\icmlauthor{Dipak Chaudhari}{rice}
\icmlauthor{Swarat Chaudhuri}{uta}
\icmlauthor{Chris Jermaine}{rice}
\end{icmlauthorlist}

\icmlaffiliation{rice}{Department of Computer Science, Rice University}
\icmlaffiliation{uta}{Department of Computer Science, University of Texas, Austin}

\icmlkeywords{Meta-learning, One-shot learning, Meta-meta classification}

\vskip 0.3in
]



\printAffiliationsAndNotice{} 

\begin{abstract}
We present a new approach, called {\em meta-meta classification}, to learning in small-data settings. In this approach, 
one uses a large set of learning problems to design an 
ensemble of learners, where each learner has high bias and low 
variance and is skilled at solving a specific type of learning 
problem. The meta-meta classifier learns how to 
examine a given learning problem and combine the various 
learners to solve the problem. 
The meta-meta learning approach is especially suited to solving few-shot learning tasks, as it is easier to learn to classify a new learning problem with little data 
than it is to apply a learning algorithm to a small data set.
We evaluate the approach on a one-shot, one-class-versus-all classification task and show that it is able to 
outperform traditional meta-learning as well as ensembling approaches.
\end{abstract}

\section{Introduction}


\textit{Meta-learning}, often defined informally as ``learning to learn'' \cite{thrun1998learning, rendell1987layered},  is a compelling approach for solving very small-data learning problems, such as one-shot or few-shot learning \cite{fei2006one}. One can generate a data set that consists of a large number of learning problems, where each problem has just a few training examples, and then use that set to learn how to solve learning problems with just a few examples.  This contrasts with competing approaches such as transfer learning \cite{torrey2010transfer}, where one solves one or more learning problems, and then adapt those solutions to a new, small-data learning problem.  Meta-learning learns the learning \emph{process}, rather than how to re-purpose an existing learner.

In this paper, we introduce a new approach to meta-learning, called \emph{meta-meta classification}.  Here, we use a large set of learning problems to design a set of $k$ different learners, each of which has high bias and low variance, so that it is skilled at solving a specific type of learning problem. Further, the meta-meta classifier also learns how to examine a new learning problem and select which of the $k$ learners should be used to solve that particular learning problem.

We call the method \emph{meta-meta classification} to distinguish it from \emph{meta-classification}, a term commonly used in ensemble methods \cite{dietterich2000ensemble}.  In ensembling, a meta-classifier is a classifier that aggregates the output from a family of learned scoring functions. For example, in bagging \cite{breiman1996bagging}, a meta-classifier may average the scores output from a family of scoring functions. In more sophisticated methods, the meta-classifier may \emph{itself} be trained so that it learns to produce an accurate output from a set of less accurate scoring functions.

In contrast, by training over a corpus of \emph{learning problems} rather than a single problem, a meta-meta classifier designs a set of learners, while at the same time learning how to examine a new problem and choose which learners are best to solve that problem. Ultimately, given a new learning problem, the output of the meta-meta classifier is a problem-specific meta-classifier defined over the set of scoring functions produced by the learners. Note that while a meta-meta classifier learns how to produce a meta-classifier, it is not itself a meta-classifier.

Meta-meta classification is particularly natural for very small-data learning problems. The underlying assumption here is that it is easier to \emph{classify} a new learning problem with little data than it is to \emph{solve} the new learning problem with little data. Intuitively, this may be the case: learned scoring functions are successfully used all the time to look at a particular object and predict its label. It does not seem to be inherently more difficult to look at a single object and its label (or small set of labeled objects in the case of few-shot learning) and identify which learners may apply to solving the problem.  If it \emph{is} possible to look at a restricted number of training examples and choose an appropriately biased, low-variance learner that best applies to the learning task, then the variance reduction realized by choosing a learner that is highly biased for the problem may result in very low error, even on highly data-restricted problems.

\vspace{5 pt}
\noindent
\textbf{Our contributions.} We define a new meta-learning strategy called \emph{meta-meta classification}, in which a meta-meta classifier is trained to recognize the type of learning task at hand, and to use that recognition to choose a biased, low-variance learner appropriate for the task.  We show how this strategy can be used to learn a highly accurate aggregate scoring function, even for one-shot learning problems.  For example, on a one-shot, one-class-versus-all classification task defined over the ImageNet corpus, meta-meta classification is able to achieve greater than 82\% test accuracy, compared to less than 61\% test accuracy for the baseline meta-learning approach, and less than 67\% for a comparatively-sized ensemble of meta-learners.

\section{Background and Problem Definition}

\subsection{Meta-Meta Classification: Overview}

\begin{figure} [t!]
  \centering
    \includegraphics[width=2.75in]{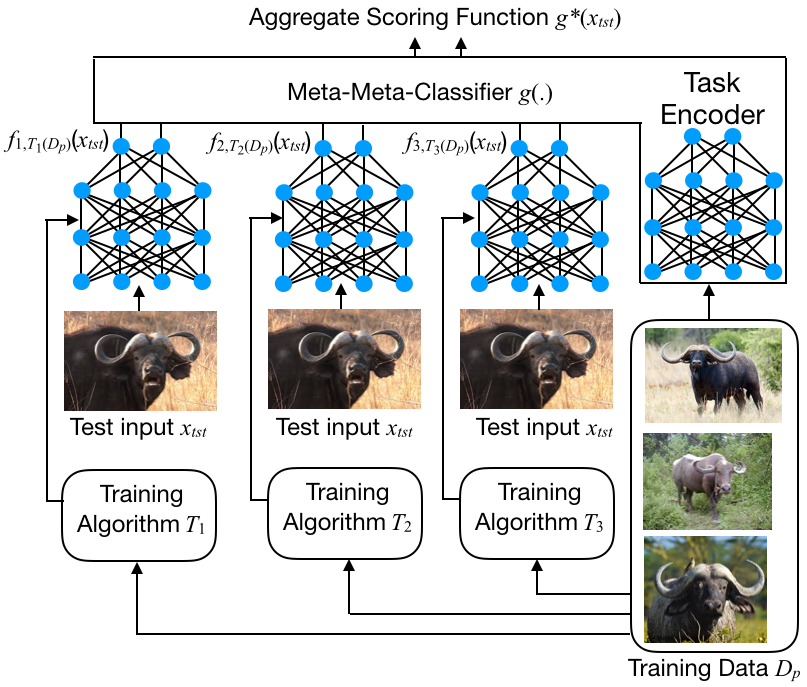}
  \caption{An aggregate scoring function realized via meta-meta classification. The meta-meta classifier $g(.)$ uses the training set $D_P$ to select from among the $k$ parameterized learners $f_{1, T_1(D_P)}$, $f_{2, T_2(D_P)}$, and so on, to realize an aggregate scoring function $g^*$.}
  \label{fig:meta-clssification}
\end{figure}

Meta-meta classification is an approach to supervised learning that is particularly relevant to the problem of one-shot or few-shot learning, as it relies on learning a set of learners designed specifically to have high inductive bias as a way to prevent over-fitting, as well as how to apply those learners when a new learning problem is encountered.

Specifically, for 
input (feature) domain $X$ and output (label) domain $Y$, 
a meta-meta classifier takes as input a training set (a multi-set drawn from $X \times Y$), and then returns an aggregate scoring function $g^{*}: X \times Y \rightarrow \mathbb{R}$ that combines the output of the learners (in the context of ensemble-based learning, this aggregate scoring function is sometimes referred to as a \emph{meta-classifier}). As in all forms of supervised learning, the goal is to produce an aggregate scoring function that gives relatively high values to pairs from $X \times Y$ that tend to occur together.

In contrast to classical ensemble approaches (such as stacking \cite{wolpert1992stacked}), in meta-meta classification, the aggregate scoring function is constructed \emph{without} examining how well the individual scoring functions output by the learners perform on the training set (or on a test set).  Instead, the meta-meta classifier learns through experience how the learners should be combined for different types of problems.  This makes meta-meta classification particularly attractive for few-shot learning problems, as there is no need to have enough data to test the accuracy of the output of the learners. 

A meta-meta classifier has two parts: a set of \emph{learners}, and a \emph{meta-aggregation function}.

\vspace{5 pt}
\noindent
\textbf{The learners.} In classical supervised learning, we have a single scoring function and a learning algorithm.  But in meta-meta classification, we instead assume an ensemble of $k$ learners, from which we wish to build an aggregate scoring function.
The $i$th learner 
consists of a scoring function $f_{i, \theta_i^f} : X \times Y \rightarrow \mathbb{R}$, as well as a training algorithm $T_{i, \theta_i^T}$. 

Let $D$ be the set of all multi-sets drawn from $X \times Y$. The training algorithm $T_{i, \theta_i^T} : D \rightarrow \Theta_i^f$ maps a set of training examples drawn from $X \times Y$ to a particular value for $\theta_i^f$.   As is typical, the scoring function $f_{i, \theta_i^f}$ is parameterized on the parameter set $\theta_i^f$ chosen from parameter space $\Theta_i^f$ by the training algorithm. More atypically, the training algorithm is itself parameterized on a parameter set $\theta_i^T$. This parameter set can contain any parameters that control the learning process: the learning rate, the number of learning iterations, the set of parameters to initialize the learning algorithm, etc. 

\vspace{5 pt}
\noindent
\textbf{The meta-aggregation function.} 
The goal is to learn, by looking at a set of learning problems, how to examine a new problem, and combine those $k$ learners to create a problem-specific meta-classifier $g^{*}$. The meta-aggregation function is given this task.

For a function $f : X_1 \times X_2 \times ... \rightarrow \mathbb{R}$, let $f(x_1, ..., x_m) : X_{m + 1}, X_{m + 2}, ... \rightarrow \mathbb{R}$ denote the function resulting from currying $f$ with respect to the first $m$ inputs, and then evaluating the resulting curried function at $(x_1, ..., x_m)$. 
Then $f_{i, T_{i, \theta_i^T}(D_{trn})}(x_{tst}) : Y \rightarrow \mathbb{R}$ is the result of applying the training algorithm in learner $i$---parameterized with $\theta_i^T$---to training set $D_{trn}$, and then ``pre-loading'' the resulting scoring function with $x_{test}$.

A \emph{meta-aggregation function} examines $D_{trn}$, and then conditioned on that $D_{trn}$, combines each of the $k$ scoring functions $f_{i, T_{i, \theta_i^T}(D_{trn})}$ to create a new, more accurate aggregate scoring function. 

Formally, a meta-aggregation function is a function:
$$g_{\theta_g} : D \times (Y \rightarrow \mathbb{R})^k \rightarrow (Y \rightarrow \mathbb{R})$$
\noindent
By allowing the meta-aggregation function to examine the set $D_{trn}$ and aggregate the scoring functions created by the $k$ learners, we obtain the aggregate scoring function $g^{*}_{\langle \theta_g, \theta_1^T,  \theta_2^T, ..., \theta_k^T \rangle} (D_{trn}, x_{tst}, y_{tst}) \equiv$ 
\begin{align}
g_{\theta_g} \biggl(D_{trn}, &f_{1, T_{1, \theta_1^T}(D_{trn})}(x_{tst}), \nonumber \\
   &f_{2, T_{2, \theta_2^T}(D_{trn})}(x_{tst}), ..., \nonumber \\
   &f_{k, T_{k, \theta_k^T}(D_{trn})}(x_{tst})\biggr) (y_{tst}). \nonumber
\end{align}
A depiction of how the learners and the meta-aggregation function together produce an aggregate scoring function $g^{*}$ is given in Figure~\ref{fig:meta-clssification}.




\subsection{Intuition: Why Meta-Meta Classification?}

If the training set $D_{trn}$ is large, it is unclear that there is much benefit to meta-meta classification. 
For large $n = |D_{trn}|$, we may choose a general-purpose learner with small inductive bias that works well regardless of the problem at hand.
However, if $n$ is small---$n = 1$ in the case of one-shot learning---there may be a significant benefit to the introduction of a set of learners and a meta-meta classifier.
If sufficient information about the problem-generating distribution $\sP$ is available through past experience, that we may learn 
a high-quality meta-meta classifier. After learning the meta-meta classifier, tiny training set $D_{trn}$ may give enough information as to the exact nature of the classification task that the meta-aggregation function can accurately select an appropriate learner. This learner will ideally have high inductive bias, and be tailored to the specific learning problem. At the same time, it will hopefully have low variance, and will be accurate, even with the learner has been trained on very small $D_{trn}$. 

In fact, this is the benefit of meta-meta classification: it allows for the use of a set of highly biased, low variance learners each of which covers a small subset of the set of classification problems that are expectedly encountered.

For this to work, a key assumption is that the task of recognizing which type of learning problem we are faced with is \emph{less data-intensive} than the task of actually solving the learning problem.  Hence, faced with limited training data, we use that data to first determine which type of learning problem we are faced with, and then use a high-bias learner that has been designed to perform well on that specific class of problem.

\subsection{Relationship to Other Approaches}

Meta-meta classification is related to several other ideas in machine learning.  For example, consider neural architecture search \cite{zoph2016neural, pham2018efficient} and related ideas. Both approaches effectively appeal to a meta-meta classifier that attempts to choose the best learner for a given task.  The key difference, however, is that
neural architecture search typically assumes large $n$, so that the meta-meta classifier is trivial. When evaluating a learner, simply see how accurate the learner is on a holdout set.  If the learned model is accurate on the holdout set, the learner is a good choice. In meta-meta classification, the assumption is that there is little data available to evaluate the accuracy of a constructed classifier, and so the meta-meta classifier $g$ is introduced as an alternative to an accuracy test over a holdout set.

There is an obvious relationship between meta-meta classification and boosting, bagging \cite{quinlan1996bagging}, and other ensemble methods. The aggregate scoring function enabled by the meta-meta classifier is effectively controlling the use of an ensemble of learners. In ensemble methods, the function that aggregates the output from an ensemble of learners is often called a meta-classifier. However, the difference is that a meta-meta classifier is trained \emph{how} to produce a task-specific meta-classifier, it is not itself a meta-classifier. By looking at a large number of learning problems, the meta-meta classifier learns how to select an appropriate, high-bias, low-variance learners from a set of learners, few of which are useful for any particular classification task.  

Meta-meta classification is related to other meta-learning approaches, for example, \cite{finn2017model}, as they also assume a distribution of learning tasks, and apply
meta-learning to try to solve the one-shot learning problem.  The key difference is that Finn et al.'s approach can be seen as 
trying to design a \emph{single} learner (scoring function plus training algorithm) that works well 
for small-sized $D_P$, for \emph{any} data-generating $P$ sampled according to $\sP$,
rather than attempting to match the present learning task with an appropriate classifier.

\section{Learning a Meta-Meta Classifier}

\subsection{Background}

Assume a universe of
probability distributions $\mathcal{P}$, each defined over the domain
$X \times Y$, and a distribution $\sP$ defined over this universe. Hence $\sP$ is a distribution \emph{of} distributions.
Now, consider the following hierarchical stochastic process for generating a triple $(D_{trn}, x_{tst}, y_{tst})$ from $\sP$: 
\begin{enumerate}
    \item Sample $P \sim \sP$
    \vspace{-5 pt} \item Sample $D_{trn} = \{(x_i, y_i)\}_{i = 1...n} \sim P$
    \vspace{-5 pt} \item Sample $(x_{tst}, y_{tst}) \sim P$
\end{enumerate}

\noindent Here, $D_{trn}$ is a training data set, and $(x_{tst}, y_{tst})$ is a test pair. 

Assume some loss function $\ell : (Y \rightarrow \mathbb{R}) \times Y \rightarrow \mathbb{R}$. That is, $\ell$ takes as an argument a scoring function defined over domain $Y$, a ``true'' value for the output selected from $Y$, and scores how accurately the scores reflect the ``true'' output. Generally, any loss function can be used for $\ell$: squared error if $Y$ is the set of real numbers, cross-entropy if $Y$ is a set of categories, etc. For example, for a scoring function $f: Y \rightarrow \mathbb{R}$, the squared error loss function is:
$$\ell_{l_2} (f, y) = \left(y - \textrm{argmax}_{\hat{y}} f(\hat{y}) \right)^2$$

The goal when learning a meta-meta classifier is to choose $\langle \theta_g, \theta_1^T, \theta_2^T, ..., \theta_k^T \rangle$ from the parameter space $\Theta_g \times \Theta_1^T \times \Theta_2^T \times ... \times \Theta_k^T$ so as to minimize the expected loss of the meta-meta classifier (or the ``meta-loss''):
\begin{align}
\mathbb{E}_{(D_{trn}, x_{tst}, y_{tst}) \sim \sP}
\left[\ell 
\left(g^{*}_{\langle \theta_g, \theta_1^T,  \theta_2^T, ..., \theta_k^T \rangle} (D_{trn}, x_{tst}), y_{tst} \right)
\right] \nonumber
\end{align}

\begin{algorithm}[t]
\caption{End-to-End Gradient Descent}
\label{alg:end-to-end}
\begin{algorithmic}
\vspace{1 pt}
\STATE \textbf{Meta-Learn} ($\sP$, $k$, $b$, $n_{trn}$, $n_{tst}$)
\STATE // $\sP$: Distribution of distributions to learn from
\STATE // $k$: \# of learners
\STATE // $b$: Meta-learning batch size (\# of problems)
\STATE // $n_{trn}$: \# of training instances in a learning problem
\STATE // $n_{tst}$: \# of test instances to evaluate a scoring function
\STATE Initialize $\theta = \langle \theta_g, \theta_1^T, \theta_2^T, ..., \theta_k^T \rangle \leftarrow \textbf{rand}()$
\WHILE {loss decreases}
  \FOR {$j = 1$ to $b$}
    \STATE Sample $P \sim \sP$
    \STATE Sample $D_{trn, j} = \{(x_i, y_i)\}_{i = 1...n_{trn}} \sim P$
    \STATE Sample $D_{tst, j} = \{(x_i, y_i)\}_{i = 1...n_{tst}} \sim P$
  \ENDFOR
  \STATE $\theta \leftarrow \theta - \frac{\alpha}{b \times n_{tst}} \sum_{j = 1}^b \sum_{(x_{tst}, y_{tst}) \in D_{tst, j}} $
  \STATE \hspace{30 pt} $\nabla \ell \left(g^{*}_{\theta} (D_{trn, j}, x_{tst}), y_{tst} \right) \left(\theta \right) \nonumber$
\ENDWHILE
\STATE \textbf{return} $\theta$
\end{algorithmic}
\end{algorithm}

There are many possible instantiations of this idea. We now briefly describe a couple of them. 

\subsection{Example: End-to-End Gradient Descent}

Assume that each of the $k$ learners utilizes gradient descent, and that $g$ is differentiable with respect to $\theta_g$.  Further, assume that $T_i$ performs one gradient update at learning rate $\lambda$ using $\theta_i^T$ as the initialization of the gradient descent, so that $\Theta_i^f = \Theta_i^T$ and:\footnote{Here, $\nabla \ell (f_{i, \theta_i^f}(x), y)(\theta_i^T)$ denotes ``the gradient of the $i$th loss function with respect to parameter set $\theta_i^f$, evaluated at $\theta_i^T$.''}
$$ T_{i, \theta_i^T}(D_{trn}) = \theta_i^T - \frac{\lambda}{n} \sum_{(x, y) \in D^T} \nabla \ell (f_{i, \theta_i^f}(x), y)(\theta_i^T).$$
\noindent Then, letting $\theta = \langle \theta_g, \theta_1^T, \theta_2^T, ..., \theta_k^T \rangle$ we can run a gradient descent algorithm to learn the meta-aggregation function parameters $\theta_g$ as well as each of the $\theta_i^T$ parameters for the various learners.  Assuming meta-learning rate $\alpha$, we repeatedly sample $(D_{trn}, x_{tst}, y_{tst}) \sim \sP$ and for each sample, apply the following update rule:
\begin{align}
\theta = \theta - \alpha \nabla \ell 
\left(g^{*}_{\theta} (D_{trn}, x_{tst}), y_{tst} 
\right) \left(\theta \right) \nonumber
\end{align}

\noindent Note that it is easily possible to extend this to training algorithms that perform more than a single gradient update; this merely requires expanding the expression computed by $T_{i, \theta_i^T}$ for an appropriate number of gradient steps. In practice, however, only a small number of gradient updates will be used in a small-data setting; a large number of steps will typically result in over-fitting.

Also, in practice, it may make sense to back-propagate the meta-loss from more than a single $(x_{tst}, y_{txt})$ test pair, as more test pairs may give a more stable estimate of the meta-loss and decrease time-until-convergence.

Finally, there is nothing preventing the use of a \emph{batch} of learning problems $P_1, P_2, ...$ during each iteration of gradient descent.  Again, this may result in a more stable algorithm that takes less time to converge.

The full algorithm for end-to-end gradient descent, which uses a batch of learning problems as well as an arbitrarily-sized test set for back-propagation is in Algorithm \ref{alg:end-to-end}.

\begin{algorithm}
\caption{Three-Step-Meta-Learning}
\label{alg:three-step}
\begin{algorithmic}
\STATE \textbf{Meta-Learn} ($\sP$, $h$, $k$, $b$, $n_{trn}$, $n_{tst}$)
\STATE // $\sP$: Distribution of distributions to learn from
\STATE // $h$: Embedding function for problem instance
\STATE // $k$: \# of learners
\STATE // $b$: Meta-learning batch size (\# of problems)
\STATE // $n_{trn}$: \# of training instances in a learning problem
\STATE // $n_{tst}$: \# of test instances to evaluate a scoring function
\STATE Initialize $\theta = \langle \theta_g, \theta_1^T, \theta_2^T, ..., \theta_k^T \rangle \leftarrow \textbf{rand}()$
\STATE \vspace{5 pt} // Cluster a set of problem instances 
\STATE $Q = \{\}$
\FOR {$m = 1$ to $big$}
  \STATE Sample $P \sim \sP$; 
  \STATE $Q = Q$ $\cup$ $\{ h(\{(x_i, y_i)\}_{i = 1...n_{trn}} \sim P) \}$
\ENDFOR
\STATE Run $k$-means on $Q$ to obtain $\mu_1, \mu_2, ..., \mu_k$
\STATE \vspace{5 pt}  // Create and partition a set of training distributions
\STATE $\sP_j = \{\}$ for $j = 1$ to $k$ 
\FOR {$m = 1$ to $big$}
  \STATE Sample $P \sim \sP$; 
  \STATE $D = \{(x_i, y_i)\}_{i = 1...n_{trn}} \sim P$
  \STATE Add $P$ to $\sP_j$ where $j =$
  $\textrm{argmin}_j$ $||\mu_j - h(D)||_2$
\ENDFOR 
\STATE \vspace{5 pt}  // Learn each of the training algorithms
\FOR {$j = 1$ to $k$}
\WHILE {loss decreases}
  \FOR {$l = 1$ to $b$}
    \STATE Sample $P \sim \sP_j$
    \STATE Sample $D_{trn, l} = \{(x_i, y_i)\}_{i = 1...n_{trn}} \sim P$
    \STATE Sample $D_{tst, l} = \{(x_i, y_i)\}_{i = 1...n_{tst}} \sim P$
  \ENDFOR
  \STATE $\theta_j^T \leftarrow \theta_j^T - \frac{\alpha}{b \times n_{tst}} \sum_{i = 1}^b \sum_{(x_{tst}, y_{tst}) \in D_{tst, i}} $
  \STATE \hspace{30 pt} $\nabla \ell \left(f_{j, T_{j, \theta_j^T}(D_{trn})}(x_{tst}, y_{tst} \right) \left(\theta_j^T \right) \nonumber$
\ENDWHILE
\ENDFOR
\STATE \vspace{5 pt} // Now, learn $g$
\WHILE {loss decreases}
  \FOR {$j = 1$ to $b$}
    \STATE Sample $P \sim \sP$
    \STATE Sample $D_{trn, j} = \{(x_i, y_i)\}_{i = 1...n_{trn}} \sim P$
    \STATE Sample $D_{tst, j} = \{(x_i, y_i)\}_{i = 1...n_{tst}} \sim P$
  \ENDFOR
  \STATE $\theta_g \leftarrow \theta_g - \frac{\alpha}{b \times n_{tst}} \sum_{j = 1}^b \sum_{(x_{tst}, y_{tst}) \in D_{tst, j}} $
  \STATE
  \vspace{-18 pt}
  \begin{align}
  \nabla \ell \biggl( g_{\theta_g} \biggl(D_{trn}, &f_{1, T_{1, \theta_1^T}(D_{trn})}(x_{tst}), \nonumber \\
   &f_{2, T_{2, \theta_2^T}(D_{trn})}(x_{tst}), ..., \nonumber \\
   &f_{k, T_{k, \theta_k^T}(D_{trn})}(x_{tst})\biggr) (y_{tst})\biggr)(\theta_g) \nonumber 
\end{align}
\vspace{-18 pt}
\ENDWHILE
\STATE \textbf{return} $\theta$
\end{algorithmic}
\end{algorithm}

\subsection{Example: Clustering Plus Gradient Descent}

\begin{figure} [t!]
  \centering
    \includegraphics[width=3.2in]{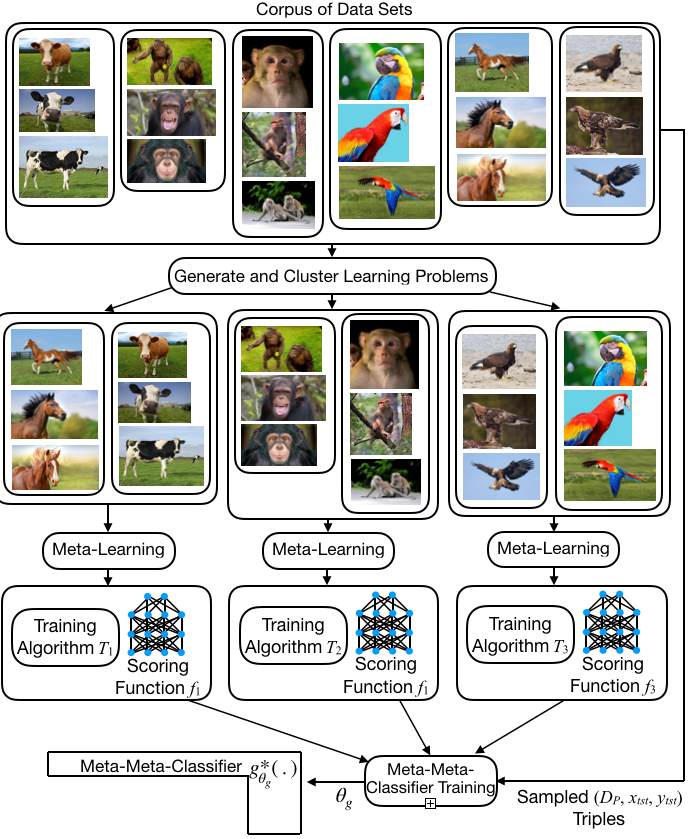}
  \caption{Learning a meta-meta classifier utilizing a pre-clustering of learning problems.}
  \label{fig:learning}
\end{figure}

Unfortunately, the algorithm from the previous subsection may not work well in practice. Note that while the meta-meta classifier is being trained in a supervised manner---the goal is to learn a meta-meta classifier that can generate an accurate meta-classifier, in one important sense, the algorithm is unsupervised. 

Ultimately, the meta-aggregation function $g_{\theta_g}$ must look at a specific training set $D_{trn}$ and determine which of the learners is most appropriate for the underlying problem. 
If, at the time that $g_{\theta_g}$ is being learned, the learners themselves are being learned, this may be viewed as an unsupervised task; it is unclear how to segment the possible problems in $\mathcal{P}$ into categories so that a reasonable learner or learners can be designed for each category. 

In practice, unsupervised learning tasks are notoriously sensitive to initialization. Few machine learning practitioners running a $k$-means algorithm would sample the initial means from a Normal$(\vec{0}, I)$ distribution, for example, as this would likely produce terrible results. Instead, the initial means may be sampled from the data set to be clustered.

Unfortunately, learning a meta-meta classifier consisting of a number of neural network learners via full gradient descent (Algorithm \ref{alg:end-to-end}), starting with a typical, random neural-network initialization for individual learning parameters $\langle \theta_1^T, \theta_2^T, ..., \theta_k^T \rangle$, is akin to initializing a $k$-means algorithm poorly. In practice, all $\theta_i^T$ values will be terrible, but one will be slightly less terrible than the others, and the meta-aggregation function will learn to route most problems to the corresponding learner. As a result, the other learners are starved of training data and ignored, and the learned solution is equivalent to what would have been returned from the MAML method \cite{finn2017model}.

One way around this is to sample a large number of distributions from $\sP$ and explicitly cluster those distributions as a separate step.  This requires having some way to cluster distributions of problems; we assume some embedding problem-specific embedding function that is able to map problem distributions (possibly non-deterministically) into a high-dimensional space, where they can be clustered using a $k$-means algorithm (here $k$ is the number of learners that are to be meta-learned). 

A procedure that uses such an explicit clustering step is depicted in Algorithm \ref{alg:three-step}. The procedure is depicted pictorially in Figure \ref{fig:learning}. After first producing the $k$ clusters of problem distributions, one leaner is meta-learned per distribution cluster.  Then, in a final step, the procedure trains the meta-aggregation function so that it is able to combine the output of the learners.

Finally, we point out that Algorithm \ref{alg:end-to-end} and Algorithm \ref{alg:three-step} can be used together. Algorithm \ref{alg:three-step} could be used to produce a high-quality initialization that is refined using Algorithm \ref{alg:end-to-end}; the combined procedure is likely to outperform either individual methodology.

\begin{table*}[t]
\centering
\caption{Experimental results. The 95\% confidence interval of observed test accuracy, computed over 10,000 problems is given. $k$ denotes the number of models trained.}
\vskip 0.15in
\begin{small}
\begin{sc}

\begin{tabular}{l|ccccr} 
\toprule
    $k$ & Whole data & Whole data & MM-classifier & \textbf{Nearest} & \textbf{Meta-meta}\\ 
      & Hard bagging & Soft bagging & on whole data & \textbf{cluster} &  \textbf{classifier} \\
      \midrule
      \multicolumn{6}{c}{ImageNet ILSVRC-2012 results}
\\
    [1ex] 
    
    2 &   61.87 $\pm$ 0.22  & 62.27 $\pm$ 0.24 & 62.79 $\pm$ 0.22 & \textbf{61.71 $\pm$  0.25}  & \textbf{66.26 $\pm$  0.20} \\

    4 & 62.48  $\pm$ 0.23  & 61.61  $\pm$ 0.24 & 63.74 $\pm$  0.23 & \textbf{69.53 $\pm$  0.22}  & \textbf{74.02  $\pm$ 0.17} \\

    8 & 62.82 $\pm$  0.24  & 62.40  $\pm$ 0.25  & 64.28 $\pm$  0.23  & \textbf{74.45 $\pm$  0.21} & \textbf{77.92 $\pm$  0.17}\\

    16 & 63.12  $\pm$ 0.24  & 63.34  $\pm$ 0.25  & 66.11 $\pm$  0.24 & \textbf{74.70  $\pm$ 0.22}  & \textbf{82.49 $\pm$  0.16}  \\ 
    \midrule
\multicolumn{6}{c}{Cross-domain results (meta-learning on ILSVRC-2012, test on CUB-2011)}
\\
    [1ex] 
    
    2 &   63.60 $\pm$ 0.22  & 64.38 $\pm$ 0.25 & 64.63 $\pm$ 0.25 & \textbf{71.53 $\pm$ 0.20}  & \textbf{70.87 $\pm$ 0.20} \\ 

    4 & 66.36 $\pm$ 0.22  & 66.27 $\pm$ 0.23 & 66.99 $\pm$ 0.23 & \textbf{69.76 $\pm$ 0.22}  & \textbf{72.44 $\pm$ 0.17} \\

    8 & 66.94 $\pm$ 0.24  & 67.04 $\pm$ 0.25  & 67.52 $\pm$ 0.25  & \textbf{74.29 $\pm$ 0.15} & \textbf{77.98 $\pm$ 0.14}\\
    
    16 & 67.21 $\pm$ 0.25  & 67.72 $\pm$ 0.26  & 69.61 $\pm$ 0.26 & \textbf{84.04 $\pm$ 0.12}  & \textbf{85.67 $\pm$ 0.11}  \\ \midrule
\multicolumn{6}{c}{Aircraft data set results}
\\
    [1ex] 

    2 &   65.39 $\pm$ 0.31  & 65.66 $\pm$ 0.30 & 69.57 $\pm$ 0.24 & \textbf{68.88 $\pm$  0.31}  & \textbf{70.65 $\pm$  0.26} \\

    4 & 70.62  $\pm$ 0.27  & 71.03  $\pm$ 0.26 & 73.00 $\pm$  0.21 & \textbf{71.72 $\pm$  0.28}  & \textbf{76.05  $\pm$ 0.23} \\

    8 & 71.84 $\pm$  0.26  & 72.23  $\pm$ 0.27  & 75.93 $\pm$  0.19  & \textbf{73.35 $\pm$  0.27} & \textbf{78.61 $\pm$  0.23}\\ \midrule
\multicolumn{6}{c}{Omniglot data set results}
\\
    [1ex] 
    
    2 &   71.24 $\pm$ 0.35  & 70.69 $\pm$ 0.29 & 73.26 $\pm$ 0.35 & \textbf{73.57 $\pm$  0.33}  & \textbf{78.70 $\pm$  0.31} \\

    4 & 73.83  $\pm$ 0.35  & 77.32  $\pm$ 0.29 & 79.16 $\pm$  0.21 & \textbf{77.07 $\pm$  0.28}  & \textbf{85.27  $\pm$ 0.18} \\

    8 & 77.70 $\pm$  0.31  & 77.61  $\pm$ 0.28  & 85.25 $\pm$  0.20  & \textbf{80.15 $\pm$  0.27} & \textbf{90.87 $\pm$  0.15}\\

    16 & 79.38  $\pm$ 0.28  & 79.56  $\pm$ 0.31  & 88.04 $\pm$  0.18 & \textbf{82.02  $\pm$ 0.26}  & \textbf{92.07 $\pm$  0.14}  \\ [1ex]

\bottomrule
\end{tabular}
\end{sc}
\end{small}

\vskip -0.1in
\label{tab:ova_results}
\end{table*}

\section{Experimental Evaluation}
\subsection{One-vs-All One-Shot Image Classification}

The first application we consider is open-world classification, where the goal is to recognize a single positive class from a large number of negative classes, some without training examples. This is one-vs-all (OvA) or one-vs-rest (OvR) classification \cite{perronnin2012towards}. Hence, we evaluate the utility of meta-meta classification for a series of one-shot image classification tasks, where the goal is to recognize---given a single example---members of a single class which are mixed in with a number of other, ``background'' classes.  We wish to answer two key questions.  First, does increasing $k$ (the number of learners) actually increase classification accuracy?  Second, does meta-meta classification outperform a simple ensemble of meta-learners?  That is, does the biased ensembling of meta-meta classification outperform the simple tactic of just using a number of independent meta-learners?

Meta-learning relies on being able to generate a distribution of learning problems.  To generate a learning problem, we sample 51 classes from the classes available for meta-learning, and one is randomly designated as a ``posi tive'' class.  The training set $D_{trn}$ is generated by sampling one image from the selected positive class, and 50 images from the 50 negative classes (some negative classes may have multiple samples, and some may not be represented in the sample set), and test set $D_{tst}$ is similarly generated by sampling 50 images from the positive class, and 50 from the negative classes.

We consider several different image classification tasks, but the first is to learn to classify images from the ImageNet database. We use the ILSVRC-2012 dataset \cite{ILSVRC15}, the most popular flavor of ImageNet data. We hold back 10\% of the 1000 ILSVRC-2012 classes for testing, and 90\% of the classes are available for meta-learning.

Each $f_i$ is the  convolutional network architecture used by \cite{finn2017model}, which has 4 modules with a 3 $\times$ 3 convolutions and 32 filters, a ReLU nonlinearity, and 2 $\times$ 2 max-pooling. The scoring function is realized using a fully connected layer after the convolutions, and the last layer is fed into a softmax.
Each $\theta_i^T$ is the initial set of weights used when training the $i$th network. During training, five iterations of gradient descent are performed.

The meta-aggregation function $g$ is realized by a simple, fully-connected neural network with two 256-neuron hidden layers. As input, this network accepts:
\begin{enumerate}
    \item $f_{i, \theta_i^f}(x_{tst}, -1)$ for $i$ in $\{1...k\}$ (that is, the ``no'' score each learner gives to the test image)
    \item $f_{i, \theta_i^f}(x_{tst}, +1)$ for $i$ in $\{1...k\}$ (the ``yes'' score that each learner gives to the test image)
    \item The 512-dimensional output of a ResNet network \cite{he2016deep}, where the final classification layers have been dropped, applied to the positive image in $D_{trn}$. This encoding allows the meta-aggregation function to classify the classification problem.
\end{enumerate}

Here, $\theta_g$ consists of the weights used in the fully-connected neural network, as well as the ResNet network used to encode $D_{trn}$.

When using the three-step training process, we sample a training set from the distribution, and our embedding function $h$ pushes the positive training instance in that set through a pre-trained ResNet network. We pre-trained a modified ResNet-152 classifier on the classes reserved for meta-learning and used the penultimate layer for feature extraction. We changed the number of output channels of the convolutions from [64, 128, 256, 512] to [64, 64, 128, 256] and block expansion from 4 to 2. This was done just to decrease the extracted feature size from the usual 2048 to 512. 

Finally, each $\theta^T_i$ is the starting parameters of the gradient descent used by the $i$th learner. Hence, in this instantiation of meta-meta classification, we are learning a set of MAML learners \cite{finn2017model}.  

\textbf{Additional One-Shot Learning Problems}.  We test three additional one-shot learning problems.

(1) Meta-learn on ImageNet ILSVRC-2012, test on the CUB-2011 Birds data set \cite{WelinderEtal2010}. In this task, meta-learning is performed exactly as above, on 900 classes selected from the ImageNet ILSVRC-2012 data set.  However, the testing distribution is different.  Each positive class for testing is selected from among the CUB-2011 Birds data set, and the negative classes are selected from among the 100 classes held back from the ILSVRC-2012 data set. The goal is to perform cross-domain testing.

(2) Meta-learn on 87 classes from the Aircraft data set \cite{maji13fine-grained}, test on 15 classes.  During testing, one of the 15 test classes is chosen as the positive class, the other 14 classes are the negative classes.  One training image is available from the positive class, and 50 from the 14 negative classes.  The goal is to perform fine-grained testing.

(3) Meta-learn on 1200 characters from the Omniglot data set \cite{lake2015human}, test on 423 characters.  During testing, a letter from the testing set is selected as the positive class, and 50 other test letters are selected as negative classes. Again, one image from the positive class is available, and 50 images of the other letters are available.

\textbf{Competitive Methods Tested}.  To evaluate the efficacy of our ideas, we compare meta-meta classification against ensembles of meta-learners.  In our experiments, the individual learners in the ensemble are MAML learners \cite{finn2017model}.  While a number of improvements to MAML have been suggested in the last couple of years (several of which are described in the Related Work section of this paper), we use MAML as a comparison point because our meta-meta classifier is effectively learning a set of MAML models. This facilitate an apples-to-apples comparison, though we note that MAML (both in meta-meta classification, and in the ensemble) could be replaced with any reasonable alternative.

Overall, we evaluate the following five classifiers: (1)
\emph{Whole-data hard bagging}: this is hard bagging over an ensemble of MAML models all trained on the entire data set. (2) \emph{Whole-data soft bagging}: soft bagging over an ensemble of MAML models. (3) \emph{Meta-meta classifier on whole data}: here we first learn a set of MAML models, each on the whole data, but then learn a meta-meta classifier (step three of three-step meta-learning) on the MAML models.  This is useful for testing the utility of segmenting the data. (4) \emph{Nearest cluster}: this is essentially the first two steps of three-step meta-learning, with the final classifier replaced with a simple nearest neighbor classifier on the ResNet features. (5) \textit{Meta-meta classifier}: this is the full three-step meta-learning.

\textbf{Results}.
For each data set and each of the five competitive methods, we test a variety of different $k$ values ($2$, $4$, $8$, $16$, though due to the small number of classes in the Aircraft data set, we omit the size-$16$ model there).  In each case, we randomly generate 10,000 learning problems to evaluate each method, and the method is scored using accuracy on 50 positive and 50 negative examples.  For the evaluation, we use five iterations of gradient descent. All results (including average accuracy, 95\% confidence interval width) are given in Table \ref{tab:ova_results}.  For comparison, a single MAML model achieved 60.78\% accruacy on ImageNet ILSVRC-2012, 62.37\% accuracy on the cross-domain bird recognition problem, and 64.92\% and 65.95\% accuracy on the aircraft and Omniglot problems.

\textbf{Discussion}.  Across all of the learning tasks, the meta-meta classifier consistently had the best accuracy---often considerably higher than the other options, and much higher than a single MAML model.  For example, on the ILSVRC-2012 data set, a meta-meta classifier with 16 classes obtains more than 82\% accuracy, compared to just under 61\% accuracy with a single MAML model.

\subsection{5-way One-Shot Image Classification}

While meta-meta classification is designed for OvA or OvR classification, it can easily be adapted to 5-way classification, which is more commonly studied in the meta-learning literature. This is done by treating a 5-way classification problem as five different OvA classification problems \cite{bishop2006pattern}. That is, given a 5-way, one-shot classification problem, we train the meta-meta classifier five times, where we cycle through the five classes, with each class in turn serving as the ``one'', and the other four classes serving as the ``all''.  Then, when it is time to classify a test image, we choose class associated with the classifier giving the highest positive score.

To test the utility of meta-meta classification for 5-way classification, we follow the procedure in \cite{finn2017model} with the ILSVRC-2012 data set. We randomly sample one image each from five randomly sampled classes as the support data for training, and randomly sample 15 images each from those classes as query data for testing. We sample 600 such problems from our test split, and measure the accuracy as well as 95\% confidence interval. On the ILSVRC-2012 data set, 16-cluster meta-meta classification obtained 57.23$\pm$1.72\% accuracy, whereas MAML gave us 49.64$\pm$1.07\% accuracy, for a net gain of +7.59\% compared to MAML. Comparing with the results presented in \cite{triantafillou2019meta} which also tested 5-way classification accuracy on ILSVRC-2012, +7.59\% net gain beats the two best meta-learning methods tested: Proto-MAML (proposed in \cite{triantafillou2019meta}; +4.02\%) and Proto-Net (proposed in \cite{snell2017prototypical}, +4.99\%). 
We also tested meta-meta classification for 5-way classification on the Aircraft data set, and obtained  57.12$\pm$1.86\% accuracy compared to MAML's 51.35$\pm$1.1\%, for a net gain of +5.77\%.

\section{Related Work}
Meta-meta classification broadly falls under the meta-learning or ``learning to learn'' paradigm \cite{hinton1987using, thrun1998learning, bengio1992optimization} which has been shown to produce promising results on few-shot classification problems. Meta-learning methods can be divided into three categories. 

First are \emph{metric-based methods} \cite{koch2015siamese, hadsell2006dimensionality, fink2005object, schroff2015facenet, shyam2017attentive, snell2017prototypical, goldberger2005neighbourhood, vinyals2016matching, taigman2015web} which aim to learn a similarity function or a distance metric between a pair of different samples. Neighborhood Components Analysis (NCA) \cite{goldberger2005neighbourhood} learns a Mahalanobis distance to maximize K-nearest-neighbors (KNN) leave-one-out accuracy. Siamese networks \cite{koch2015siamese} use a pairwise verification loss to perform nearest-neighbours classification. Matching Networks \cite{vinyals2016matching} combine both embedding and classification to form an end-to-end differentiable nearest neighbours classifier. Prototypical Networks \cite{snell2017prototypical} apply an inductive bias in the form of class prototypes without full context embeddings.

Second are \emph{memory-augmented methods} \cite{munkhdalai2017meta, mishra2017simple, duan2016rl, wang2018prefrontal, santoro2016meta, oreshkin2018tadam} that learn to adjust model states using memory-augmented recurrent networks. For example, \cite{santoro2016meta} represents entries from a sample set in an external memory, AdaResNet \cite{munkhdalai2017rapid} uses memory and the sample set to produce conditionally shifted neuron coefficients for the query set, and SNAIL \cite{mishra2017simple} uses an explicit attention mechanism to leverage specific information from past experience.

Third are \emph{optimization based methods} \cite{finn2017model, finn2018probabilistic, yoon2018bayesian, lee2018gradient, grant2018recasting, nichol2018reptile, rusu2018meta, rothfuss2018promp, ravi2016optimization, zhang2018metagan} that learn a network initialization that can quickly adapt to new tasks within a distribution of tasks with a very few steps of regular gradient descent. MAML \cite{finn2017model} backpropagates the meta-loss through an inner learning loop, Reptile \cite{nichol2018reptile} incorporates an L2 loss that updates the meta-model parameters towards the task-specific models, and \cite{lee2018gradient} learns a layer-wise subspace where gradient-based adaptation is done. However, since all of these meta-learners sample a task from a task-distribution to learn the initial parameters, they can be prone to overfitting \cite{mishra2017simple}. 


\section{Conclusion}

We have explored a new type of meta-learning, called \emph{meta-meta classification}.  The idea in meta-meta classification is to learn a set of learners tailored to different problem types, as well as a function called a ``meta-meta classifier'' that is able to look at a particular problem and decide how to combine the learners to solve that problem.  Thus, a meta-meta classifier itself meta-learns to produce a meta-classifier over the output of the learners. Meta-meta classification is predicated on the assumption that it is easier to classify a problem (and choose an appropriate set of learners) than it is to learn to solve the problem with little data. We have shown through a series of experiments that meta-meta classification can have much higher accuracy than a standard meta-learner or even an ensamble of such meta-learners.

\newpage
\balance
\bibliography{mmc}
\bibliographystyle{icml2020}

\newpage
\section*{Appendix}
\beginsupplement

\subsection{Broader Impact}

One-shot classification is an important but difficult problem, with many applications in science and technology.  For example, consider the problem of searching a database of brain images, to find images with a particular type of brain lesion, where only a few (or one) images of the desired injury are available. Our proposed method could easily be used in this case. Note that this example application is an example of ``one-vs-all, open-world, one-shot classification'', which is under-studied (or never-before studied), and which we consider in this paper.  Almost all prior work considers $n$-way classification (for $n$ = 5).  Our new method performs very well on this problem, compared to the obvious alternatives. For another particular application, consider the task of recognizing one animal species (for which only one positive image is available) from among a set of images taken by an automated wildlife camera. A set of negative training images may be available, but the set of negative classes cannot be controlled. Or, consider recognizing a particular vessel (boat) from among a large set of satellite images of vessels in the open ocean. In our opinion, these are as meaningful as the more common 5-way classification task seen in most few-shot learning papers. Our work (and future work) on this sort of one-vs-all problem could greatly increase the range of problems amenable to one-shot classification.
However, as with any classification algorithm, especially in small data regime, our meta-meta classifier runs the risk of misclassification in critical applications, like the above example of recognizing a particular type of brain lesion. In those cases, it is more important to measure the precision and/or recall (depending upon the problem) rather than the classification accuracy and use a confidence interval to apply our model with caution.

\subsection{Data Sets}

In our experiments, we have used four publicly available data sets. A short description of these data sets is given below.

\begin{figure}[hbt!]
\centering
\subfigure[ILSVRC-2012]{%
\includegraphics[width=0.2\textwidth]{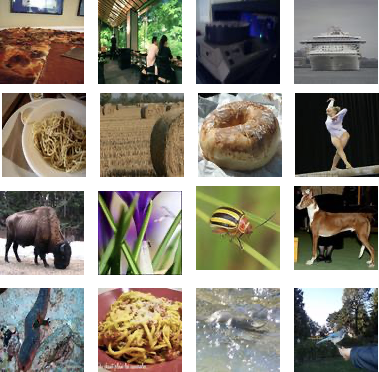}
\label{fig:ilsvrc}}
\quad
\subfigure[CUB-2011 Birds]{%
\includegraphics[width=0.2\textwidth]{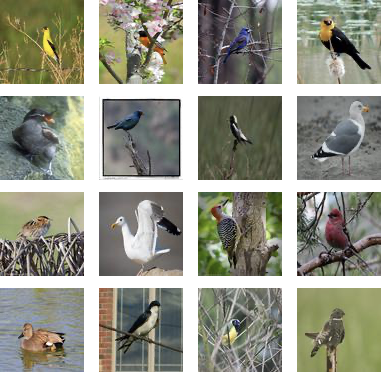}
\label{fig:cub}}
\subfigure[Omniglot]{%
\includegraphics[width=0.2\textwidth]{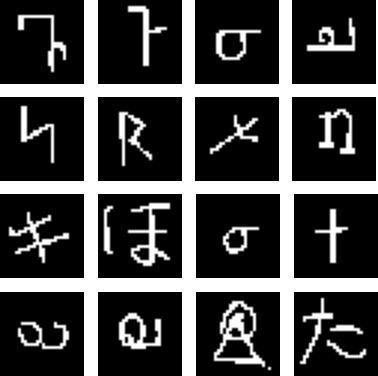}
\label{fig:omniglot}}
\quad
\subfigure[Aircraft]{%
\includegraphics[width=0.2\textwidth]{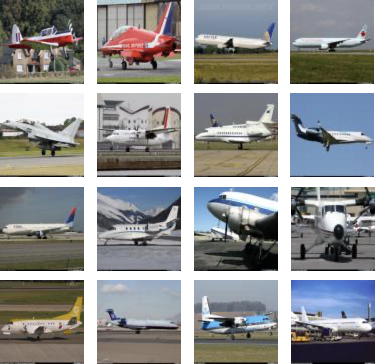}
\label{fig:aircraft}}
\caption{Training examples sampled from the data sets used in our experiments.}
\label{fig:dataset}
\end{figure}

\textbf{ILSVRC-2012.} \cite{ILSVRC15} A data set of natural images of 1000 diverse categories, the most commonly used Imagenet data set, primarily released for `Large Scale Visual Recognition Challenge' (Figure A\ref{fig:ilsvrc}). Randomly chosen 900 classes are used for training and the remaining 100 classes are used for testing.

\textbf{CUB-200-2011.} \cite{WelinderEtal2010} A data set for fine-grained classification of 200 different bird species, an extended version of the CUB-200 dataset (Figure A\ref{fig:cub}). All 200 classes are used for cross-domain testing.

\textbf{Omniglot.} \cite{lake2015human} A data set of images of 1623 handwritten characters from 50 different alphabets, with 20 examples per class (Figure A\ref{fig:omniglot}). Randomly chosen 1200 characters are used for training and 423 classes are used for testing.

\textbf{Aircraft.} \cite{maji13fine-grained} A dataset of images of aircrafts spanning 102 model variants, with 100 images per class (Figure A\ref{fig:aircraft}). Randomly chosen 87 classes are used for training and 15 classes are used for testing.

Note that in each data set, the validation set is included in the training split.

\textbf{Data Processing.} All images apart from those from Omniglot are resized into 84x84 resolution. Omniglot images are resized into 28x28 resolution and color-inverted for faster training.

\subsection{Sampling one-shot problem instances}
Let the training split of $n_{mtrn}$ classes for meta-training be called $S_{mtrn}$ and test split of $n_{mtst}$ classes for meta-testing be called $S_{mtst}$. Meta-learning relies on a distribution of learning problems which are sampled as follows.

\noindent\textbf{Meta-training}. 
We sample one class from $S_{mtrn}$ and designate it as a ``positive'' class, called $C^{<pos>}_{mtrn}$. Then we sample 50 ``negative'' classes, called $C^{<neg>}_{mtrn, trn}$, from the remaining ($n_{mtrn}$-1) classes of $S_{mtrn}$. Now, the support set or training set, $D_{trn}$, is generated by sampling one image from the selected positive class, and 50 images from the 50 negative classes. The sampling from negative classes is done with replacement, so some negative classes may have multiple samples, and some may not be represented in $D_{trn}$. For the query set or test set, $D_{tst}$, we use the same positive class and again sample 50 ``negative'' classes, $C^{<neg>}_{mtrn, tst}$, from the remaining ($n_{mtrn}$-1) classes of $S_{mtrn}$. Then $D_{tst}$ is generated by sampling 50 images from the positive class, and 50 images from the negative classes with replacement. Note that $C^{<pos>}_{mtrn}$ is a single class whereas $C^{<neg>}_{mtrn, trn}$ and $C^{<neg>}_{mtrn, tst}$ are sets of 50 classes.

So, in total, we sample 151 images which constitute one learning problem for meta-training as follows:

\noindent $D_{trn}$: 1 from $C^{<pos>}_{mtrn}$, 50 from $C^{<neg>}_{mtrn, trn}$ \\
$D_{tst}$: 50 from $C^{<pos>}_{mtrn}$, 50 from $C^{<neg>}_{mtrn, tst}$.

\noindent\textbf{Meta-testing}. Sampling learning problems for meta-testing is exactly the same as for meta-training, only difference being we use $S_{mtst}$ instead of $S_{mtrn}$ data. So, here also, we sample 151 images which constitute one learning problem for meta-testing as follows:

\noindent $D_{trn}$: 1 from $C^{<pos>}_{mtst}$, 50 from $C^{<neg>}_{mtst, trn}$ \\
$D_{tst}$: 50 from $C^{<pos>}_{mtst}$, 50 from $C^{<neg>}_{mtst, tst}$.

Note that when we run MAML on clusters, for each cluster, the positive image comes from that particular cluster and all other images come from the entire meta-training set, $S_{mtrn}$.

\subsection{Architecture and Hyperparameters}

For pretraining a feature extractor, we use a modified ResNet-152 \cite{he2016deep} model where we change the number of output channels of the convolutions from [64, 128, 256, 512] to [64, 64, 128, 256] and block expansion from 4 to 2. For MAML, we use the commonly-used four-layer convolutional network \cite{finn2017model}. While we acknowledge that the meta-learners have shown improved performance with resnet-18 architecture \cite{triantafillou2019meta}, we keep the basic four-layer convnet for consistency in our method as well as the comparing methods. Since we design our method as a binary classifier, we avoid batch-normalization. This is because while doing the testing during meta-learning, batch-normalization can learn to distinguish the negative images from the positive ones without training, provided negative images come from multiple negative classes.

In order to realize the meta-aggregation function, we use a fully connected feed forward network with two hidden layers of 256 units. The input to the meta-aggregator is the concatenation of the 512 dimensional pretrained ResNet-152 feature vector of the `positive' support image and the binary logits of the corresponding `negative' query images coming from individual learners. We use dropout with probability 0.9 for the input layer and 0.6 for the hidden layers during training.

Finally, we tune the learning rate schedule and weight decay and we use ADAM optimizer to train all the models. Other details and the complete set of hyperparameters used are included in the source code.


\end{document}